\documentclass[runningheads,a4paper]{llncs}

\usepackage{amssymb}
\usepackage{graphicx}

\usepackage{url}

\begin{document}

\mainmatter  % start of an individual contribution

% first the title is needed
\title{Analysis of Cancer Omics Data In A Semantic Web Framework}
% a short form should be given in case it is too long for the running head
\titlerunning{Semantic Web Reasoning on Translational Data}

% the name(s) of the author(s) follow(s) next
%
% NB: Chinese authors should write their first names(s) in front of
% their surnames. This ensures that the names appear correctly in
% the running heads and the author index.
%
\author{Matthew E. Holford\inst{1}
\and James P. McCusker\inst{2} \and Kei-Hoi Cheung\inst{3,4,5} \and Michael Krauthammer\inst{1,2}}
\authorrunning{M. Holford et al.  }
% (feature abused for this document to repeat the title also on left hand pages)

% the affiliations are given next; don't give your e-mail address
% unless you accept that it will be published
\institute{Interdepartmental Program in Computational Biology \& Bioinformatics,
\and
Department of Pathology, 
\and
Department of Computer Science,
\and
Center for Medical Informatics
\and
Department of Genetics
\\ 
Yale University\\
New Haven, CT
}

%
% NB: a more complex sample for affiliations and the mapping to the
% corresponding authors can be found in the file "llncs.dem"
% (search for the string "\mainmatter" where a contribution starts).
% "llncs.dem" accompanies the document class "llncs.cls".
%

\maketitle

\begin{abstract}
  Our work concerns the elucidation of the cancer (epi)genome,
  transcriptome and proteome to better understand the complex
  interplay between a cancer cell's molecular state and its response
  to anti-cancer therapy.  To study the problem, we have previously
  focused on data warehousing technologies and statistical data
  integration.  In this paper, we present recent work on extending our
  analytical capabilities using Semantic Web technology.  A key new
  component presented here is a SPARQL endpoint to our existing data
  warehouse.  This endpoint allows the merging of observed
  quantitative data with existing data from semantic knowledge sources
  such as Gene Ontology (GO).  We show how such variegated
  quantitative and functional data can be integrated and accessed in a
  universal manner using Semantic Web tools.  We also demonstrate how
  Description Logic (DL) reasoning can be used to infer previously
  unstated conclusions from existing knowledge bases.  As proof of
  concept, we illustrate the ability of our setup to answer complex
  queries on resistance of cancer cells to Decitabine, a demethylating
  agent.
\end{abstract}

\section{Introduction}
The Yale Specialized Program in Research Excellence (SPORE) in skin
cancer is a large translational cancer project, which aims at rapidly
moving biological insights from the ``bench to bedside''. As part of
the effort, the SPORE collects skin cancer samples from mostly
malignant melanoma patients and performs a multitude of Omics studies,
probing the melanoma genome, epigenome, transcriptome and
proteome. The idea is to integrate this data with clinical outcome
information to derive prognostic and predictive biomarkers,
i.e. genomic markers that predict patient survival and drug therapy
effectiveness, respectively. Conventionally, these markers are either
derived statistically in an unbiased fashion \cite{van2002}, or by
prior knowledge and candidate (gene) selection \cite{koga2009}. We are
interested in combining these approaches, and are developing means for
unbiased assessment of Omics data using existing knowledge on cellular
processes that affect drug effectiveness. In particular, we are
employing Semantic Web technology to create RDF graphs that define the
genomic state of cancer cells and the functional annotation of the
cells' molecular entities (i.e. genes or proteins). We use SPARQL to
query these graphs to better understand the molecular basis of drug
resistance and sensitivity.

We start by retrieving quantitative data from a large relational
database, a component of the Corvus architecture \cite{mccusker2009},
storing melanoma Omics data.  We do this by providing a new semantic
component of Corvus, a SPARQL endpoint which relies upon Hibernate
\footnote[1]{http://www.hibernate.org} for Object Relational Mapping
(ORM).  Through this endpoint, we can dynamically create RDF graphs of
the data stored within.  We then merge such graphs with SKOS-converted
Gene Ontology (GO) \cite{ashburner2000} information to annotate
genomic elements with functional data, such as their involvement in
certain cellular processes.

As a case study, we used the new Corvus SPARQL endpoint to create an
RDF graph with data representing drug response to Decitabine, a
demethylating agent that has been shown to be clinically active in
melanoma \cite{jabbour2008}.  Using SPARQL, we queried Corvus for
melanoma samples with information on promoter methylation status and
gene expression before and after Decitabine treatment.  The resulting
graph is augmented with functional annotations from GO.  It is then
interrogated for the molecular mechanisms explaining why some samples
have better response to Decitabine treatment than others.

\section{Methods}
To attain these goals, we needed to build a model that integrated
quantitative Omics data with functional information.  Our model
incorporates gene expression and methylation data for seven melanoma
cell lines \cite{halaban2009}; it also contains Gene Ontology (GO)
annotations for the whole of the human genome.  Expressing this model
as an RDF triple store affords us a number of advantages.  First, it
provides a way for others to borrow from and build upon our work.  It
allows us to use the standardized SPARQL interface to perform queries
that bridge quantitative and functional knowledge.  It also gives us
the capability to infer previously unstated information by reasoning
over the data with a Semantic Web aware Description Logic (DL)
reasoner.  We attempted wherever possible to borrow terms from
well-established OBO ontologies \cite{smith2007}.  Doing so places our
work under the auspices of community defined best practice and allows
our model to be used in conjunction with similarly designed knowledge
bases.  Building the model involved the use of a variety of
cutting-edge Semantic Web technologies and required the creation of
several novel tools.  The work proceeded along two major lines:
(i). Conversion of relational data from melanoma cell lines to RDF/OWL
and (ii). Integration of specific gene annotations with the Gene
Ontology.

The issue of integrating quantitative and functional biological
information to infer relevant new information has been frequently
explored.  A notable example is HyBrow, a tool for the generation and
evaluation of biological hypotheses \cite{racunas2004}.  The user can
derive hypotheses from HyBrow's knowledge base of functional
biological information and test them against various high-throughput
data sources.  BioBIKE offers an environment for users to integrate a
wide variety of experimental and genomic data to reach new conclusions
\cite{elhai2009}.  Originally released as a LISP interactive library
\cite{massar2005}, the software is now web-based to accommodate users
lacking in programming expertise.  When combined with the BioDeducta
module, it enables automated reasoning \cite{shrager2007}.  Although
both HyBrow and BioBIKE make extensive use of ontologies, neither is
Semantic Web enabled.  Recent efforts by the National Cancer Institute
as part of the caBIG initiative \cite{fenstermacher2006} have focused
on addressing the integration issue though the use of an
Extraction-Transform-Load (ETL) strategy.  Notably, the caIntegrator2
\footnote[2]{http://cabig.nci.nih.gov/tools/caIntegrator2} project
uses ETL to integrate quantitative Omics data from caArray
\cite{heiskanen2005} and functional biological data from caBio
\cite{covitz2003}.  The Bio2RDF project is notable for providing
normalized URIs for a wealth of identifiers and relationships from
functional biology in the hopes of allowing easier integration of
diverse data sets \cite{belleau2008}.

\subsection{Quantitative Data From Melanoma Cell Lines}
We examined data derived from seven melanoma cell lines (WW165, YUMAC,
YUGEN8, YUSAC2, YUSIT1, YULAC and YURIF).  These lines have been
experimentally classified using IC50 values from dose-response
analysis as being either sensitive to (YUMAC, YUSAC2, YULAC, YUSIT1,
YUGEN8) or resistant to (WW165, YURIF) decitabine
(5-Aza-2'-deoxy-cytidine, Aza), a DNA methyltransferase inhibitor.
Specifically we looked at relative methylation values prior to
administration of AZA and the ratio of gene expression following
administration of AZA to before.  The methylation values were obtained
from a Nimblegen promoter array using the Methyl-DNA
immunoprecipitation (MeDIP) technique \cite{paik2004,pelizzola2008}.
Gene expression ratios were obtained using a custom 2-channel
Nimblegen array.  Data from both arrays are available for download
through ArrayExpress \footnote[3]{http://www.ebi.ac.uk/arrayexpress/}.
We used the Gene Element Ontology (GELO) to align the array probes to
RefSeq identifiers \cite{szpakowski2009}.

\subsubsection{Rationale for building a SPARQL endpoint}
These data were stored in a relational database component of Corvus, a
data warehouse for experimental data, which currently holds over 4
million observations from diverse Omics experiments across melanoma
cell lines.  Presently, Corvus exists as a Java library with
object-relational mapping (ORM) accomplished through Hibernate.
Quantitative cancer omics data is stored in a standard database schema
specified by the ORM.  We present here a new semantic interface to
Corvus which retrieves data in the form of RDF triples.
Unfortunately, the sheer volume of data contained within our local
Corvus database would result in a triple store of such size as to be
untenable for the purposes of DL reasoning.  What was needed instead
was a way to retrieve a subset of the Corvus model containing only the
information essential to the problem at hand.  Ideally this could be
accomplished in a dynamic fashion.  

Integration of traditional relational databases with RDF has been
extensively explored in recent years \cite{sahoo2009}.  Typically the
approach is to create a generic mapping between relational and RDF
schema.  This has been done either through automatic mappings, where
relational tables correspond to RDFS classes and relational columns to
RDF predicates \cite{chen2004}, or with domain-specific semantics
\cite{sahoo2008}.  Some tools, such as d2rq \cite{bizer2004}, provide
for both and allow user customization for complex cases such as when
mappings are not one-to-one.  Mappings may be stored in a variety of
formats, ranging from XML configuration files to custom languages such
as R2O \cite{barrasa2004}.  These mapping artifacts can then be used
to dynamically generate SQL queries to the relational database based
upon queries expressed according to the RDF schema, usually using
SPARQL.

We experimented directly with the d2rq framework, which allows a
relational database to be queried like a triple store using SPARQL.
Using a configuration file to map Corvus database fields to RDF
properties, we were able to generate SPARQL queries that retrieved a
manageable subset of the Corvus database.  However, we found that the
SQL generated by the tool to query the relational database was
inefficient and data retrieval took longer than expected.  We decided
instead to leverage the Hibernate mappings already part of the Corvus
model to interact with the relational database.  We wrote a SPARQL
interface to the Corvus model which interacts directly with the Java
library, taking advantage of Hibernate's ability to optimize and cache
relational queries.  To the best of our knowledge, although the issue
of mapping SPARQL to object oriented representations such as Hibernate
has been discussed \cite{corno2008,hillairet2009}, no tools for doing
this have been released to the public.  Our approach is to create
wrapper classes around the Hibernate mapping classes which map the
property getters to RDF predicates.  Indirect mappings make possible
situations in which the RDF and relational schemas do not correspond
one to one.  Though this approach is not necessarily a universal
solution, we felt that given Corvus' ability to represent such a broad
swathe of Omics data, the performance gain offered by these customized
mappings more than justified the up-front expense of their creation.

\subsubsection{Corvus model to RDF mapping}
\begin{figure}
  \includegraphics[width=\textwidth]{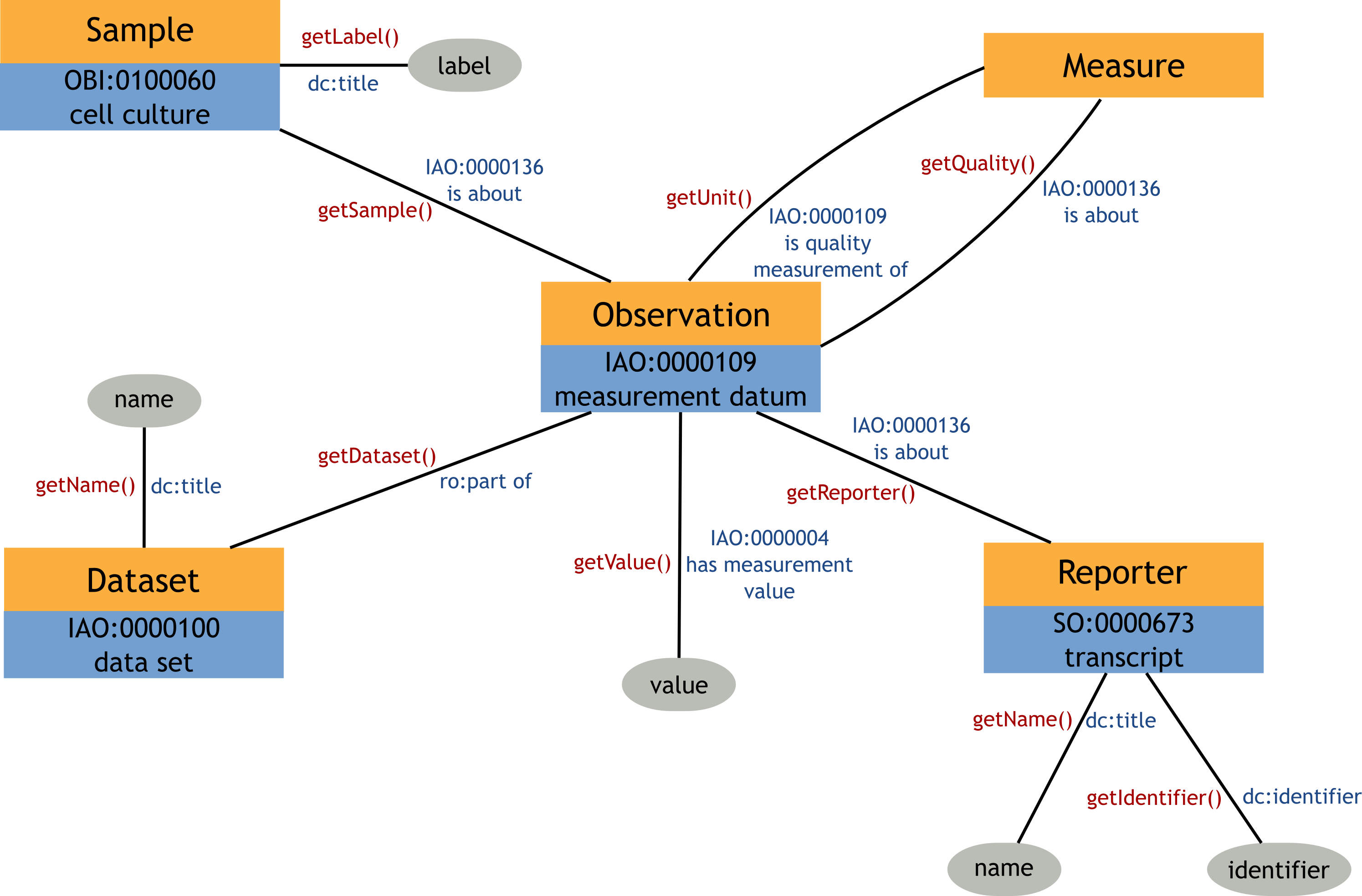}
  \caption{Diagram showing Java classes in the Corvus model (orange
    boxes) next to their corresponding OWL classes (blue boxes).  Data
    or annotation properties are shown as gray ellipses.  Edge labels
    show the Java method used to call the Corvus model in red and the
    RDF property used in the semantic model in blue.}
  \label{fig:rdfschema}
\end{figure}

We mapped fields from the Corvus database to classes and relationships
from OBO ontologies.  In particular, we employed terms from
Information Artifact Ontology (IAO)
\footnote[4]{http://code.google.com/p/information-artifact-ontology}
and Ontology for Biomedical Investigations (OBI) \cite{courtot2009}.
In addition to being actively developed, these ontologies are notable
for building upon the foundation Basic Formal Ontology (BFO)
\footnote[5]{http://www.ifomis.org/bfo} and the OBO Relation Ontology
(RO) \cite{smith2005} which were specially designed to be extensible
by any biomedical ontology.  This allows our modeled Corvus data to be
incorporated with other OBO ontologies with relative ease.  It should
be noted that we are simply borrowing terms from these ontologies, not
incorporating them in their entirety as doing so would have a
significant deleterious effect on reasoning performance.  This does
not pose a hindrance to our goals as we do not need to make inferences
across the whole hierarchy of terms in these ontologies.  By using the
terms, however, we provide an entry point for others who may wish to
explore this type of inferencing in the future.  Quantitative data
storage in the Corvus model is centered around the
\textit{Observation} class.  Instances of this class represent
individual data points in a collection of data, such as an array.
They contain the numerical value of the data as well as pointers to
other classes indicating the type and provenance of the data.  These
other classes include \textit{Dataset}, which holds metadata on
experimental conditions, \textit{Measure}, which specifies details
about the type of data being measured, \textit{Sample}, which
describes the cell line being measured and \textit{Reporter}, the
genomic feature (typically a gene) for which data is being reported.
We mapped \textit{Measurement} to the IAO class \textit{measurement
  datum} and used the IAO data property \textit{has measurement value}
to associate numerical data values.  \textit{Dataset} was linked to
the IAO class \textit{data set}.  Individual \textit{Observation}s can
be specified as belonging to a \textit{Dataset} using the RO property
\textit{part of}.  Samples were declared as instances of the OBI class
\textit{cell culture}.  Association of an \textit{Observation} with a
\textit{Sample} was done using IAO's \textit{is about} property.
\textit{Reporter} was linked to the \textit{Genomic Region} class from
the GELO ontology.  This class is defined as a superclass of the OBO
Sequence Ontology's (SO) \cite{eilbeck2005} \textit{biological region}
class and it attaches properties to assign a reference location for a
genomic element within the genome.  For the purposes of our data,
\textit{Reporter}s were made instances of SO's \textit{transcript}
class, as the reference sequence (RefSeq) was used.  We used the
Uniform Resource Identifiers (URIs) for RefSeq sequences provided by
the Bio2RDF project.  Using this normalized identifier allows us to
easily link with other resources describing the same genes.  To
capture information from the Corvus \textit{Measure} class, instead of
mapping to an instance of a class, we forwarded two of
\textit{Measure}'s fields to properties in the domain of
\textit{measurement datum}.  These were the IAO \textit{is quality
  measurement of} property and the IAO \textit{has measurement unit
  label} property.  Finally, we used the Dublin Core \cite{weibel1997}
annotation properties \textit{title} and \textit{identifier} to assign
names for \textit{Sample}s, \textit{Dataset}s and \textit{Reporters}
and reference identifiers to \textit{Reporters}.  A detailed view of
this model is provided in figure \ref{fig:rdfschema}.

\subsubsection{Querying the Corvus SPARQL endpoint}
To retrieve a subset of our Corvus database that was sufficient for
our ultimate querying purposes, we issued a SPARQL query that would
retrieve all relevant information for the seven cell lines mentioned
above.  We used a SPARQL \textit{DESCRIBE} query which simply returns
all relevant properties for a type into an RDF graph.  Our query
retrieves all \textit{Observation}s associated with the cell lines and
pulls in information on the lines and experimental conditions from the
\textit{Sample} and \textit{Dataset} tables and all genes with values
from the \textit{Reporter} table.  We issued the following SPARQL
query for each of the seven cell lines:
\begin{verbatim}
PREFIX obo: <http://purl/obolibrary.org/obo/>
PREFIX dc: <http://purl.org/dc/elements/1.1/>
PREFIX ro: <http://www.obofoundry.org/ro/ro.owl#>

DESCRIBE ?rep ?obs ?data ?samp
WHERE {
  ?samp dc:title ``YUMAC'' .
  # IAO_0000136 = 'is_about'
  ?obs obo:IAO_0000136 ?samp .
  ?obs ro:part_of ?data .
  ?obs obo:IAO_0000136 ?rep .
}
\end{verbatim}
Retrieval of a populated RDF graph containing the approximately
120,000 observation for a cell line using our Hibernate-based mapping
typically took between one and two minutes.

\subsection{ Annotated GO Terms}
To include functional information about genes, we decided to
incorporate the well-known Gene Ontology (GO).  GO is presented in the
OBO format, a simple model for expressing hierarchies of terms and the
relationships between them.  Although significantly less powerful for
inferencing than a fully DL-compatible language like OWL, the OBO
language makes it straightforward to declare relationships between
classes of object.  We found an effective compromise to be the use of
the Simple Knowledge Organization System (SKOS) \cite{miles2005}.  In
this ontology, written in OWL, terms such as those in OBO taxonomies
are expressed as instances of a \textit{Concept} class.  Class
subsumption is handled though OWL object properties that describe
\textit{Concepts} as \textit{broader} or \textit{narrower} than other
\textit{Concept}s.  In this system, properties can be assigned easily
to class-like terms without violating the strictures of OWL-DL.  This
approach offers significant advantages for querying and reasoning, as
the common alternative, creation of restrictions on classes, is
computationally expensive while still requiring the creation of
individual instantiations to infer properties.  Using the OBO to SKOS
conversion tools developed at University of Manchester
\footnote[6]{http://www.cs.man.ac.uk/~sjupp/skos}, we created a
GO-SKOS ontology which converts GO terms to instances of
\textit{Concept} and \textit{is a} relationships to \textit{broader}
relationships.

\begin{figure}
  \includegraphics[totalheight=0.7\textheight]{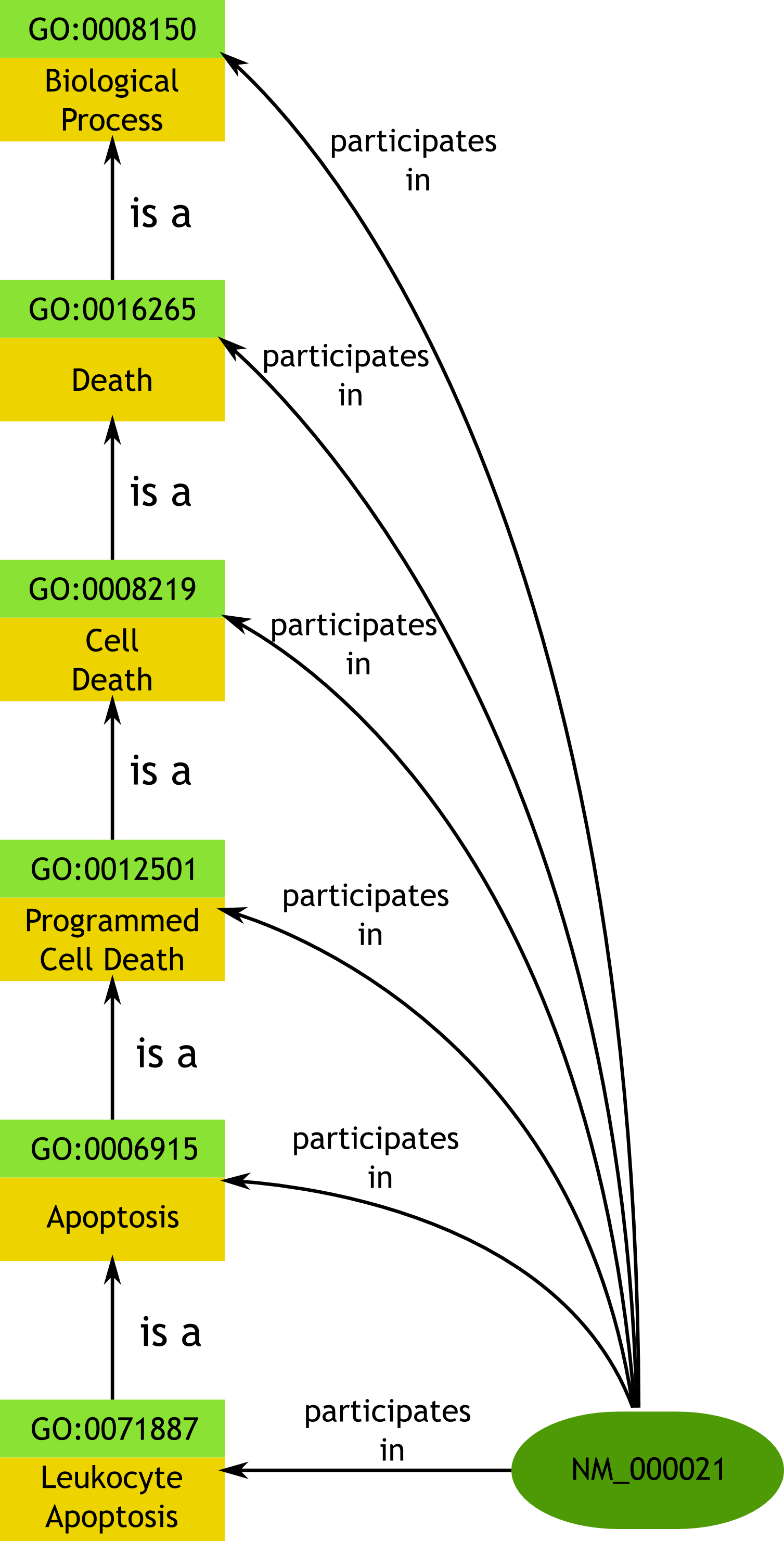}
  \caption{Diagram showing the propagation of the participates in
    property up the class subsumption hierarchy.  This inference is
    achieved by using an OWL 2 property chain associating the
    participates in property with the SKOS broader property.}
  \label{fig:goa}
\end{figure}

We downloaded the standard human genome annotations provided by the
Gene Ontology consortium.  In order to easily merge with our Corvus
graph, we converted the GO annotation file's HUGO symbols to RefSeq
identifiers using conversion tables made available from Entrez
\footnote[7]{http://www.ncbi.nlm.nih.gov/Entrez} and used the Bio2RDF
normalized URIs.  In fitting with the Corvus model, we cast individual
refseqs as instances of the SO:transcript class.  We then used three
basic relationships from RO to link the gene to its appropriate term
in whichever of GO's three main hierarchies.  Genes annotated with a
Biological Process term were linked using \textit{participates in};
those labeled as expressing a Molecular Function were linked using
\textit{has function} and genes marked as being located in a
particular Cellular Component were linked using \textit{part of}.  We
also wished for the properties assigned to genes to propagate up the
chain of hierarchy.  In other words, if a particular gene participates
in a specific biological process, we wanted the reasoner to be able to
infer that it also participates in the more generic process.  For
example, genes participating in apoptosis also participate in the more
general process of cell death and in biological processes in general.
To accomplish this, we used an OWL property chain, a new feature in
OWL 2, to associate \textit{participates in} with \textit{broader},
stating that if A participates in B and C is a broader concept than B,
then A participates in C as well.  This type of inference is possible
because the is-a (subsumption) relationship between SKOS concepts is a
relationship between individuals rather than between classes.  The
relationship is illustrated in figure \ref{fig:goa}.  We made the same
declarations for the \textit{has function} and \textit{part of}
properties.

With these declarations in place we were able to run the ontology
through a DL reasoner and create a greatly expanded set of RDF triples
with all inferences spelled out (i.e. all annotation properties
propagated along the hierarchy).  There is a trade-off here as we gain
faster query times by precomputing all inferences at the expense of
additional storage space and less flexibility, as we need to recompile
when the underlying data changes.  Creation of the fully entailed GO
annotation RDF graph took approximately five minutes on our Linux
workstation using 8 GB of memory.

\subsubsection{Merging of RDF graphs}
\begin{figure}
  \includegraphics[width=\textwidth]{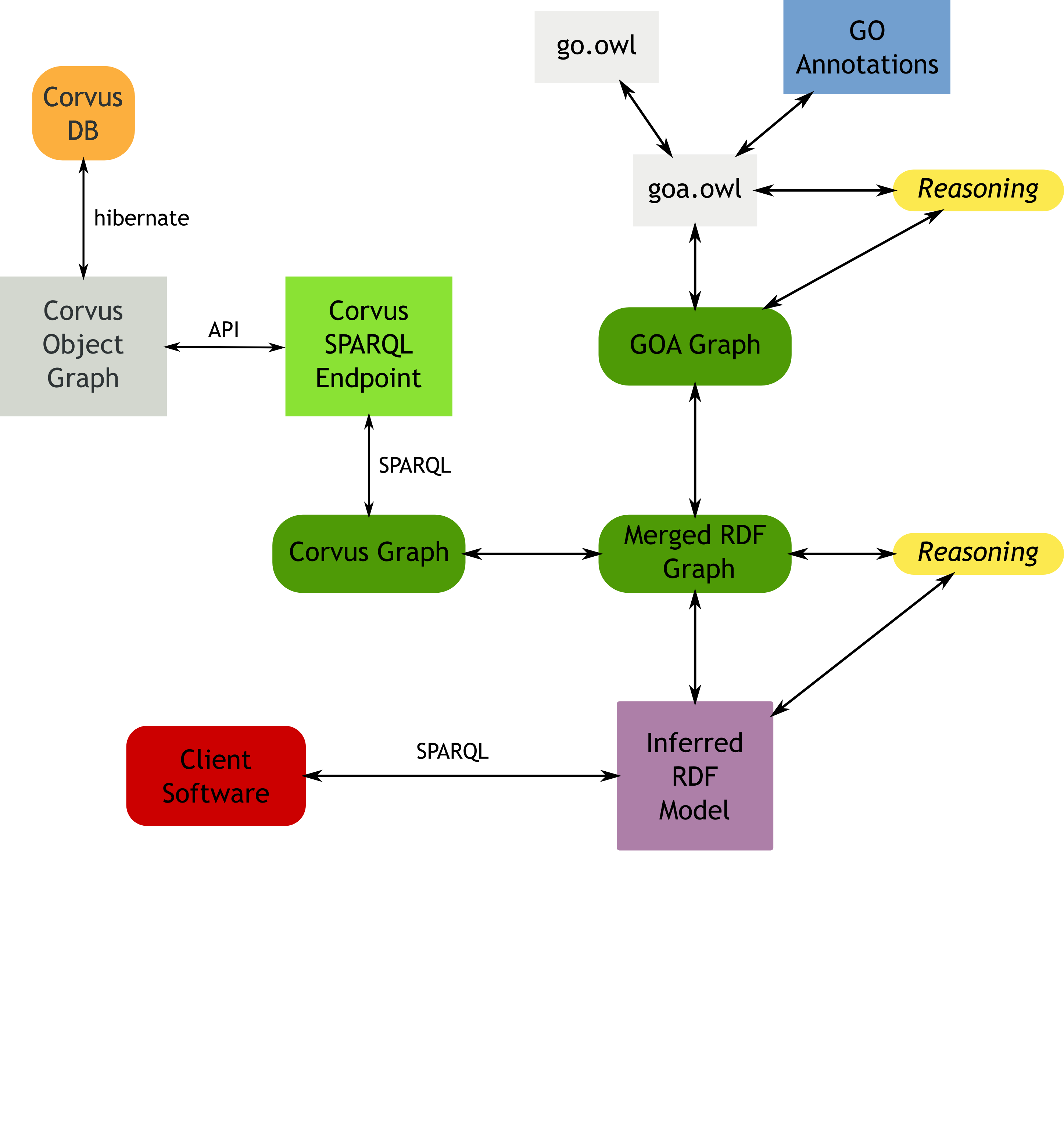}
  \caption{Diagram showing the architecture of the integrated model we
    used to perform the queries in this paper.}
  \label{fig:architecture}
\end{figure}

The GO annotation model could at this point be merged with the Corvus
quantitative data model, the points in common being the instances of
SO \textit{transcript} representing individual RefSeqs/genes.  Because
we use identical URIs from the Bio2RDF namespace to describe these
instances, we can assure that we are referring to the same gene in the
two sources.  This merged model could now be queried using SPARQL.
The full architecture of our setup for creating an RDF graph from
Corvus and merging it with the GO graph is shown in figure
\ref{fig:architecture}.

The Corvus SPARQL endpoint Application Programming Interface (API) was
written in Java making extensive use of the Jena API for RDF
manipulation and the closely related ARQ API for SPARQL processing
\footnote[8]{http://jena.sourceforge.net}.  The GO Annotation
pre-processing was handled by a Java program making use of the OWLAPI
OWL2 library \cite{bechhofer2003} and the Pellet DL reasoner for
Semantic Web data \cite{sirin2007}.  Merging of the ontologies was
also handled by Java code using first the ARQ API to issue the SPARQL
query on the relational Corvus store and then OWLAPI to perform the
actual merge.  The merged dataset was loaded into an instance of TDB,
an RDF triple store employing the Jena libraries.  It was then loaded
into a running instance of Joseki, a web application allowing
execution of SPARQL queries over HTTP.  Joseki also uses the Jena
libraries extensively.  An endpoint for the merged dataset is
available at \url{http://doppio.med.yale.edu:2020/sparql}.

\section{Results and Discussion}
We wanted to show that it was possible to use Corvus to execute
arbitrarily complex queries incorporating information across varied
knowledge domains.  To this end, we tried to verify cell lines that
were resistant or sensitive to Decitabine, a demethylating agent used
for melanoma therapy.  Our formulated query asks for genes involved in
apoptosis with high methylation values prior to Decitabine
administration and increased gene expression following.  We use values
from two datasets obtained from the Corvus SPARQL endpoint, relative
methylation values prior to treatment and ratio of gene expression
post- to pre-treatment.  Apoptosis-related genes were found using the
merged triples from the GO annotations.  Our SPARQL query was as
follows:
\begin{verbatim}
PREFIX dc: <http://purl.org/dc/elements/1.1/>
PREFIX ro: <http://www.obofoundry.org/ro/ro.owl#>
PREFIX obo: <http://purl.obolibrary.org/obo/>
PREFIX go: <http://purl.org/obo/owl/GO#>

SELECT distinct ?rep ?samp
WHERE {
      ?ds dc:title "Methylation Relative" .
      ?obs ro:part_of ?ds .
      # IAO_0000004 = 'has_measurement_value'
      ?obs obo:IAO_0000004 ?obsVal .
      # IAO_0000136 = 'is_about'
      ?obs obo:IAO_0000136 ?rep .
      ?obs obo:IAO_0000136 ?samp .
      # OBI_0100060 = 'cell celture'
      ?samp a obo:OBI_0100060 .
      ?ds2 dc:title "AZA Pre-Post Treatment Ratios" .
      ?obs2 ro:part_of ?ds2 .
      ?obs2 obo:IAO_0000136 ?rep .
      ?obs2 obo:IAO_0000136 ?samp .
      ?obs2 obo:IAO_0000004 ?obsVal2 .
      ?rep ro:participates_in go:0006915 .
      FILTER ( ?obsVal > 2  ) .
      FILTER ( ?obsVal2 > 1 )
} 
\end{verbatim}
This query returns the URIs of genes and cell lines that match the
aforementioned criteria.  Using features from the recently
standardized SPARQL 1.1, we can aggregate genes by cell line to get a
count of highly expressed genes per cell line.  The slightly modified SPARQL query is: 
\begin{verbatim}
PREFIX dc: <http://purl.org/dc/elements/1.1/>
PREFIX ro: <http://www.obofoundry.org/ro/ro.owl#>
PREFIX obo: <http://purl.obolibrary.org/obo/>
PREFIX go: <http://purl.org/obo/owl/GO#>

SELECT (count(?rep) as ?repcount) ?samp
WHERE {
      ?ds dc:title "Methylation Relative" .
      ?obs ro:part_of ?ds .
      # IAO_0000004 = 'has_measurement_value'
      ?obs obo:IAO_0000004 ?obsVal .
      # IAO_0000136 = 'is_about'
      ?obs obo:IAO_0000136 ?rep .
      ?obs obo:IAO_0000136 ?samp .
      # OBI_0100060 = 'cell celture'
      ?samp a obo:OBI_0100060 .
      ?ds2 dc:title "AZA Pre-Post Treatment Ratios" .
      ?obs2 ro:part_of ?ds2 .
      ?obs2 obo:IAO_0000136 ?rep .
      ?obs2 obo:IAO_0000136 ?samp .
      ?obs2 obo:IAO_0000004 ?obsVal2 .
      ?rep ro:participates_in go:0006915 .
      FILTER ( ?obsVal > 2  ) .
      FILTER ( ?obsVal2 > 1 )
} GROUP BY (?samp)
\end{verbatim}

We can compare these counts to what we know from experimental data
regarding the level of sensitivity/resistance of various cell lines
\cite{halaban2009}.  The results are shown in the following table:
\begin{figure}
  \centering
  \begin{tabular}{|l|c|c|}
    \hline
    Cell Line & Gene Count & IC50 (nM)\\
    \hline
    YUMAC & 22 & 34\\
    YUSAC & 7 & 91 \\
    YULAC & 9 & 110 \\
    YUSIT1 & 2 & 132 \\
    YUGEN8 & 6 & 139 \\
    WW165 & 2 & 239 \\
    YURIF & 0 & 255\\
    \hline
  \end{tabular}
  \caption{Table showing the seven melanoma cell lines, the total
    number of apoptosis-related genes positively expressed that were
    formerly methylated and the IC50 value.}
  \label{tbl:qresults}
\end{figure}
The sensitive cell lines with low IC50 values (YUMAC, YUSAC and YULAC)
had the three highest gene counts, whereas the two most resistant
lines (WW165 and YURIF) had the lowest.  As the mechanism of
Decitabine action is demethylation of gene promoters, and
(re)expression of the corresponding genes, these results give rise to
the following hypothesis: Decitabine targets apoptosis-related gene
promoters predominantly in Decitabine-sensitive cell lines, thus
conveying its cytotoxic effect by activating the apoptosis pathway.
The following validation steps are warranted to strengthen the
hypothesis: First, one might want to independently test in vitro both
the demethylation of the implicated gene promoters, as well as the
re-expression of the corresponding genes.  Also, the finding should be
repeated in a larger cohort of melanoma samples.  A current limitation
of our SPARQL query is that we only interrogate for fold change after
Decitabine treatment.  As shown in prior work, the absolute change in
expression values after treatment should also be taken into account
\cite{rubinstein2010}.

\section{Conclusion}

Our proof of concept query illustrates how easily data from various
sources can be integrated using the common framework of OWL/RDF.  It
reveals some of the power of Semantic Web reasoning and querying tools
for inferring and elucidating discovered knowledge.  It also shows the
importance of customization in mapping non-semantic data to RDF.
While generic tools mapping relational data to RDF have recently
emerged, our experience with d2rq has shown that there are still areas
where direct mapping is significantly more efficient and flexible.
Our work also makes a strong case for the benefits of using linked
data, as use of the Bio2RDF normalized URI for RefSeqs made
integration of the two branches of our ontology a breeze.

The flexibility of the Corvus model will allow us to incorporate
quantitative Omics data from a variety of modalities.  In the future,
this could include cancer data from caArray or caIntegrator or data
obtained directly from ArrayExpress using MAGETab2RDF
\cite{mccusker2010}.  Essentially, Corvus functions as a
contextualized observation repository and we intend to incorporate
information from other contexts including clinical data and generic
provenance data.  We hope to use the new semantic access point to
Corvus to integrate this data with other types of information such as
pathway and pharmacological data.  The simplicity and elegance of the
integrated Semantic Web approach also suggests its usefulness as an
access point to making sense of variegated data for researchers
unequipped with the programming or mathematical expertise to work with
traditional data mining tools.

\subsubsection*{Acknowledgments.} 
This work has been supported by the National Cancer Institute (Yale
SPORE in skin cancer - 5P50CA121974) and the National Library of
Medicine (Yale Biomedical Informatics Research Training Program -
5T15LM007056).

\bibliographystyle{splncs03}
\bibliography{swats}

\end{document}